\RequirePackage{fix-cm}
\documentclass[smallextended]{svjour3}       % onecolumn (second format)
\smartqed  % flush right qed marks, e.g. at end of proof
\usepackage{graphicx}
\usepackage{epstopdf}
\usepackage{amssymb}
\usepackage{graphicx}
\usepackage{algorithm}
\usepackage{algorithmic}
\usepackage{amsmath}
\usepackage{amsmath}
\usepackage{amssymb}
\DeclareMathOperator*{\argmax}{argmax}
\usepackage{tikz}
\usepackage{pgfplots}
\usepackage{filecontents}
\usepackage{float}
\usepackage{array}
\usepackage{graphics}
\usepackage{multirow}
\usepackage{tabularx}
\usepackage{bbding}
\usepackage{url}
\usepackage{indentfirst}
\usepackage{cite}
\usepackage{enumerate}

\usepackage{fancyhdr}
\begin{document}
\title{Generalized Zero-Shot Learning for Action Recognition with Web-Scale Video Data
}
\author{Kun Liu  \and Wu Liu  \and Huadong Ma \and Wenbing Huang \and Xiongxiong Dong }


\institute{ Kun Liu \and Wu Liu  \and Huadong Ma \and Xiongxiong Dong \at
              Beijing University of Posts and Telecommunications, Beijing, China \\
              Tel.: +086-62-282767\\
              Fax: +086-62-282767\\
              \email{\{liu\_kun, liuwu, mhd, dong-bupt\}@bupt.edu.cn}
           \and
           Wenbing Huang \at
              Tencent AI Lab, Beijing, China \\
             \email{helendhuang@tencent.com}
   }
\date{Received: date / Accepted: date}
\maketitle
\begin{abstract}
Action recognition in surveillance video makes our life safer by detecting the criminal events or predicting violent emergencies. However, efficient action recognition is not free of difficulty. First, there are so many action classes in daily life that we cannot pre-define all possible action classes beforehand. Moreover, it is very hard to collect real-word videos for certain particular actions such as steal and street fight due to legal restrictions and privacy protection. These challenges make existing data-driven recognition methods insufficient to attain desired performance. Zero-shot learning is potential to be applied to solve these issues since it can perform classification without positive example. Nevertheless, current zero-shot learning algorithms have been studied under the unreasonable setting where seen classes are absent during the testing phase.
Motivated by this, we study the task of action recognition in surveillance video under a more realistic \emph{generalized zero-shot setting}, where testing data contains both seen and unseen classes. To our best knowledge, this is the first work to study video action recognition under the generalized zero-shot setting. We firstly perform extensive empirical studies on several existing zero-shot leaning approaches under this new setting on a web-scale video data. Our experimental results demonstrate that, under the generalize setting, typical zero-shot learning methods are no longer effective for the dataset we applied. Then, we propose a method for action recognition by deploying generalized zero-shot learning, which transfers the knowledge of web video to detect the anomalous actions in surveillance videos. To verify the effectiveness of our proposed method, we further construct a new surveillance video dataset consisting of nine action classes related to the public safety situation.
%Third, we propose a novel method for action recognition in surveillance video by deploying generalized zero-shot learning, which transfers the knowledge of web video to detect the anomalous actions in surveillance videos.


\end{abstract}

\section{Introduction}
\label{intro}
In recent years, action recognition in surveillance video has drawn considerable attention \cite{cristani2007audio,luo2016action,georgakopoulos2007event} from the computer vision research community owing to its potential for public safety. Meanwhile, the application of surveillance cameras is dramatically growing due to the declining cost for semi-conductor components. Obviously, the huge amount of surveillance cameras produces massive surveillance video data at the same time. Therefore how to automatically and efficiently recognize semantic actions from them becomes an urgent problem to be solved. For example, if we are able to recognize human action in surveillance video, such as criminal events or violent emergencies, the terrorist attacks will be blocked and many lives will be saved. We thus can make our city smarter and safer through employing action recognition algorithms in the intelligent video surveillance, which can be applied to public safety, personal health care, traffic management, resource allocation, etc.

However, action recognition in surveillance video also faces many great challenges. First, video is a complicated modality which is mainly composed of several crucial and complementary cues: static appearances, temporal dynamics, and audio information.
On top of the challenges that make object class recognition hard, there are a number of complexities, like camera motion, scale variations,
and the continuously changing viewpoints that come with it.
Whereas convolutional neural networks (CNNs) \cite{krizhevsky2012imagenet} have achieved great success for image classification, some attempts \cite{wang2015action,wang2016compact,gan2015devnet} are made to encode the video level information by aggregating the individual frame features. Therefore, a two-stream CNN is proposed in \cite{simonyan2014two} which cohorts spatial and temporal net by capturing single RGB image appearance and calculating optical flow to explicitly capture motion information respectively.
The two-stream architecture and its improved algorithms \cite{diba2016deep,girdhar2017actionvlad} exhibited promising performance on standard web-scale video datasets \cite{kuehne2013hmdb51,soomro2012ucf101} for action recognition.

%In addition, the high complexity of the video also bring considerable computational cost for action recognition, thus making it a bottleneck for current methods applied to large-scale datasets. A real-time two-stream network is proposed in \cite{zhang2016real} by replacing the optical flow with the motion vector to skip the computationally expensive step, namely the calculation of optical flow. Transferring the knowledge learned with optical flow CNN to motion vector CNN can tackle the problem of evident degradation of recognition performance caused by inaccurate motion patterns in the motion vector. Similarly, Luo et al.\cite{luo2016action} design a new motion feature representation named motion history image to train the temporal net, which can achieve impressive trade off between accuracy and recognition speed.

Moreover, newly defined action concepts come out everyday so that the action categories occurred in surveillance video keeps updating all the time. This leads to the fact that we cannot define all possible action labels beforehand. To solve this issue, some paradigms focus on training a series of attribute classifiers and then recognize the target actions based on the attribute representation of action category. More precisely, the Direct Attribute Prediction(DAP) method \cite{lampert2014attribute} learns probabilistic attribute classifiers by calculating the class posteriors and then predicting the class label through MAximum a Posterior (MAP) estimate.

The last but not least great challenge for action recognition especially in the surveillance video is the lack of sufficient training data. Existing action recognition methods require many positive exemplars to train classifiers for each action. However, different from the web-scale video datasets \cite{karpathy2014large,kuehne2013hmdb51,soomro2012ucf101,gao2016constrained} which can be constructed by unrestrainedly downloading video from internet and annotating them with extensive labor, a large-scale dataset in surveillance video is hard to be built as we have no access to collect enough surveillance videos through public method. Moreover, collecting the videos which record dangerous events or illegal activities is more difficult due to legal restrictions and privacy protection. The limited videos for detecting abnormal actions makes it hard to train current methods especially the deep neural based ones.

To the best of our knowledge, zero-shot learning \cite{changpinyo2016synthesized,gan2016concepts,gan2016learning} is a promising alternative approach to solve above issues because it can recognize new categories when testing classes do not have any training sample available. Zero-shot learning differentiates two types of classes: seen and unseen, where only instances in seen classes have available labels. During training phrase, for seen classes, both the visual information and semantic labels are available; for unseen classes, only semantic labels of the unseen classes are available without any visual labelled instance. In generally, a map function projected seen classes to unseen classes is learned to construct the models for unseen classes. This is usually achieved by embedding both seen and unseen classes into a common semantic space.
For example, the one-hot vectors can be built by judging the category whether has property visual attributes.
Moreover, Word2Vec \cite{mikolov2013efficient} representations provide a distributed representation of categories by
using a learned word embedding for the word associated with each category.
This common semantic space can transfer models from the seen categories to the unseen ones.
As illustrated in Fig.~\ref{fig:01},
in our example of recognizing smashing car and street fight without any positive videos,
the attribute-based representation requires to collect attribute information of these actions such as ``street fight is outdoor relate'' or ``smashing car has stick-like tool''. Word2Vec representations of the class names also can be adopted to implement the projection from seen categories to the unseen ones.

However, the existing zero-shot learning methods have been studied in an unrealistic setting---\emph{the testing set only consists of unseen classes}.
This setting for zero-shot learning is that once models for unseen classes are learned, they are
tested only on unseen classes, hypothesizing no seen categories during the testing phase.
This assumption does not truly reflect the real-world action recogniton situations.
In practice, the seen actions are always more common than the unseen classes and testing set is always from both seen classes and unseen classes.
\emph{Therefore, in this paper, we investigate the generalized zero-shot learning task, where the testing set contains both seen and unseen classes.}
Fig.~\ref{fig:01} presents the differences of zero-shot and generalized zero-shot learning tasks.
More precisely, under the zero-shot learning setting, the learned model is evaluated only on unseen
classes (street fight and smash car) at testing time. In contrast, under generalized zero-shot setting, the testing task is required to predict both seen and unseen classes (i.e., street fight, smash car, do karate, and remove ice from car, all of which are likely appeared in real life). It is unreasonable to assume that the seen data is absent during the testing stage.
A truly useful approach should not only accurately discriminates among either seen or unseen categories themselves but also exactly judges among both the seen and unseen ones.
Without a doubt, the generalized zero-shot learning is more difficult as the trained models are more likely to classify the videos into the seen classes.

The contributions of this paper are summarized as follows.
\begin{itemize}
\item We take efforts to build a surveillance video dataset which consists of nine important action categories for public safety. The training examples of the nine categories are very hard to be collected. This dataset is used to evaluate the generalized zero-shot learning problem and will be publicly available.
\item This paper is the first work to recognize actions in the generalized zero-shot learning setting and it provides a baseline for this task. We do extensive experiments by applying several zero-shot learning approaches to the generalized setting on a standard web video benchmark and our built dataset.
\item We transfer the knowledge of existing large-scale web video dataset to learn classifiers for actions occurred in surveillance video through zero-shot learning algorithm. The comprehensive evaluation on our dataset demonstrates that the proposed method is a promising direction for solving the problem of action recognition in surveillance video.
\end{itemize}

The rest of the paper is organized as follows. In the next section,
we briefly review the related works on zero-shot learning and action recognition.
We introduce our approach in Section \ref{ourme}. Experimental results and detailed analysis on ActivityNet \cite{caba2015activitynet} and our own dataset are provided in Section \ref{exper}. Finally, Section \ref{concl} concludes the paper.

\begin{figure}[t]
\begin{minipage}[b]{0.95\linewidth}
  \centering
  \centerline{\includegraphics[width=12cm,height=5.2cm]{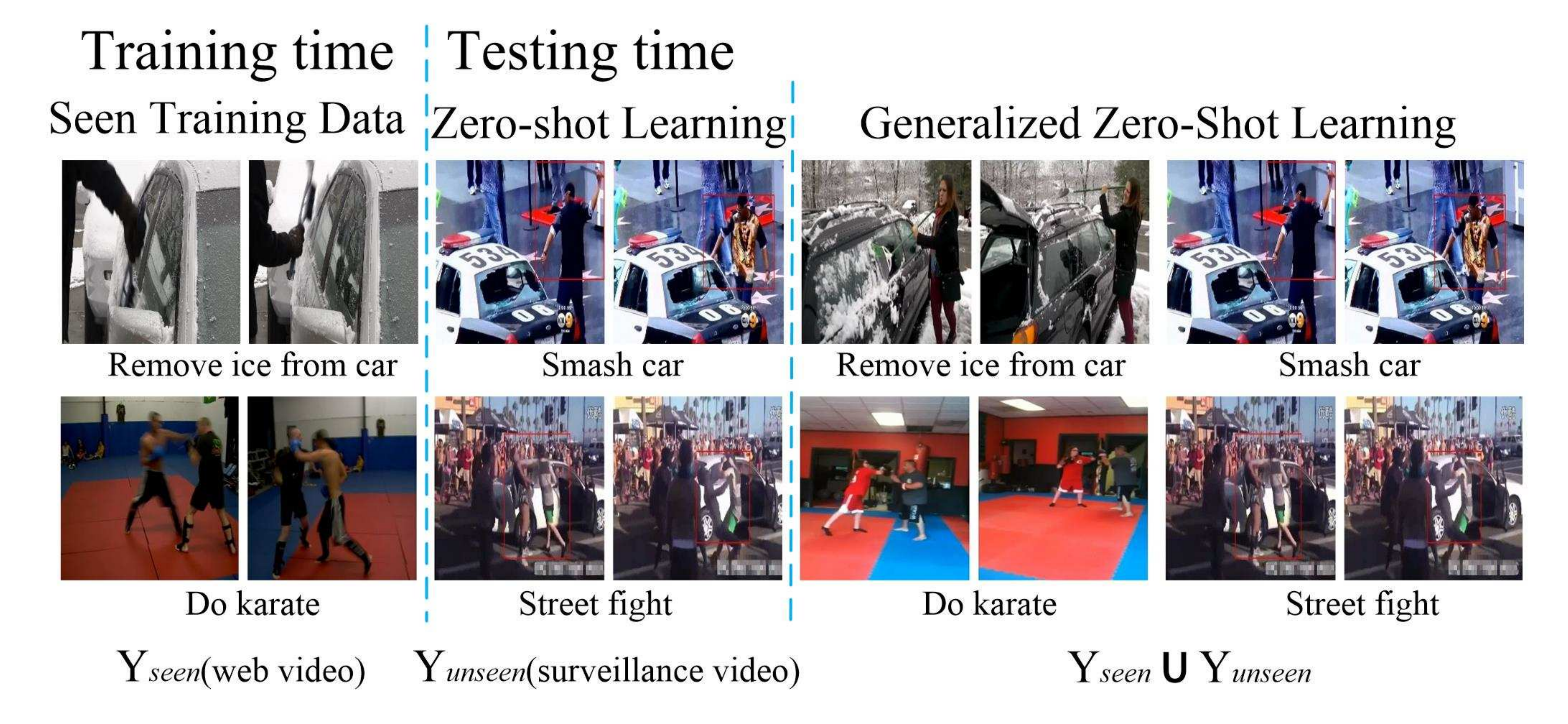}}
  \label{fig:01}
\end{minipage}
\caption{ Demonstration of the differences between zero-shot learning and generalized zero-shot learning.  At training time, both the visual information and semantic spaces of the seen classes ($Y_s$) are available. During testing time, at the zero-shot learning setting, the learned model is judged only on unseen classes ($Y_u$). Differently, at generalized zero-shot setting, new data is classified into both seen and unseen label spaces. (Best seen in color and with zoom.)}
\label{fig:01}
\end{figure}

\section{Related Works}
\label{relat}
\indent This paper involves two research directions: zero-shot learning and action recognition, which are reviewed in this section.

\subsection{Action Recognition.}
Video-based action recognition has been widely explored in the past few years. Previous works closely related to ours fall into two types: (1) Action recognition with hand-crafted features; (2) CNN based Action Recognition.
In the early stage, some action recognition techniques focus on designing powerful and effective video representations using local spatio-temporal features, such as Motion Boundary Histograms(MBH) \cite{dalal2006human}, 3D Scale-Invariant Feature Transform (SIFT-3D) \cite{scovanner20073} and Histogram of Optical Flow (HOF) \cite{laptev2008learning}.
For example, Wang et al. \cite{wang2013action} recently propose a state-of-the-art hand-crafted feature named improved Dense Trajectories (iDT), which extracts several descriptors (HOG, HOF and MBH) along the trajectory.
Then the feature distributions are encoded by the robust residual encoders such as Vector of Locally Aggregated Descriptors (VLAD) and its probabilistic version Fisher Vector \cite{arandjelovic2013all}. In the final step, a classifier such as Support Vector Machine (SVM) is learned for classification. In spite of superior performance on various datasets,
this kind of methods not only lack discriminative capacity as well as scalability, but also is computationally intensive so that it is difficult to be applied to large scale datasets.

Recently some efforts \cite{diba2016deep,girdhar2017actionvlad,simonyan2014two,karpathy2014large,gan2016recognizing} have been made to go beyond individual image static appearance feature and exploit the motion information based on CNN.
Simonyan and Zisserman \cite{simonyan2014two} propose a two stream architecture which consists of spatial and temporal CNN. The input of spatial and temporal network is individual RGB frame and stack of ten dense optical flow maps, respectively. This could be the first work demonstrating that the deep model is more accurate than hand-engineered features, such as IDTs-based representation.
After it, many algorithms are proposed to improve the two stream networks from different aspects.
For example, Wang et al. \cite{wang2015action} develop an effective video presentation, called trajectory-pooled deep-convolutional descriptor (TDD), which shares the merits of hand-crafted features and deep-learned features.
Feichtenhofer et al. \cite{feichtenhofer2016convolutional} propose a novel architecture which can incorporate the appearance and motion information to make full use of spatio-temporal information by fusing spatial and temporal network.
Moreover, Wang et al. \cite{wang2016temporal} propose a temporal segment network to encode the long-range temporal structure over the whole video by feeding the network with several segments from one video.
A real-time two stream network is proposed in \cite{zhang2016real} which replaces the optical flow with motion vector to skip computationally expensive step.
Tran et al. \cite{tran2015learning} explore deep 3-dimensional convolutional network to learn spatiotemporal features trained on Sports-1M \cite{karpathy2014large}, which is a extreme large-scale web video dataset containing one million video examples across around five hundred categories.
Experiments on public datasets demonstrate that this 3D architecture can model appearance and motion information simultaneously. The learned features are compact, efficient, and extremely simple to be used.

It's worth noting that most of the above methods achieve promising performance based on the large scale web-scale video datasets \cite{kuehne2013hmdb51,soomro2012ucf101}, which include several hundred or thousand training examples per action class. The lack of training data in the surveillance video directly leads to the sharp degradation of recognition performance. Since zero-shot learning is capable of recognizing action categories without ever having seen them before, we deploy zero-shot learning algorithms to recognize the events occurred in surveillance video even there is no available instance to train classifiers.

\subsection{Zero-Shot Learning.}
The task of zero-shot learning is to recognize classes that have never been seen before. Namely, there are no positive exemplars available to train a set of action classifiers.
Zero-shot learning is demanded to have the ability to transfer knowledge from seen classes to unseen classes where we do not have corresponding training data by making use of other forms of information about unseen classes, such as attributes or visual class hierarchy.
Then a semantic embedding of classes or an attributes-to-classes mapping can be obtained to transfer the knowledge from seen to unseen classes.

For example, Liu et al. \cite{liu2011recognizing} propose a piece of well-structured attribute list to recognize actions, which is probably the first attempt to recognize actions only using texts. The attributes are comprised of human comprehensible properties that are shared and reused across different classes, such as translation motion or
torso up-down motion.
As presented in \cite{liu2011recognizing}, the ability of characterizing actions by attributes is not only helpful for recognizing available actions and events, but also powerful for classifying novel action classes for which no training samples are available.
In conclusion, the majority of recent methods yield the attributes by 1) manual annotation, 2) extracting the features themselves, or 3) mining knowledge from other domains. After acquiring the attributes, the effectiveness of knowledge transferring always relies on the performances of trained classifiers independently or the mapping function between low-level features and attribute labels. Besides learning intermediate attribute classifiers,
there are many works focusing on learning a mixture of seen class proportions or by a direct approach, e.g. compatibility learning frameworks.
ConSE \cite{norouzi2013zero} is one of the most widely used representatives of learning a mixture of class proportions. ALE \cite{akata2016label} and DEVISE \cite{frome2013devise} both propose frameworks to tackle the problem of absence of training sample by learning compatibility. These two methods have different loss functions.
ESZSL\cite{romera2015embarrassingly} further studies this problem and proposes to add a regularization term to unregularized compatibility learning methods.
CMT\cite{socher2013zero} extends the compatibility learning algorithm by proposing non-linear extensions to the framework.

However, zero-shot learning is analyzed under a hypothesis that the instances of seen classes is absent from the testing set. In the real application, this setting might be impractical since it is usual to encounter instances in both seen and unseen classes during the testing phase.
To better investigate this problem, Chao et al. \cite{chao2016empirical} advocate a generalized zero-shot learning setting, where models trained on samples of seen classes require to predict testing data from both seen and unseen classes. Xian et al. \cite{xian2017zero} first reconstruct some popular benchmarks for fair comparison and then analyze a significant number of the state-of-the-art zero-shot learning methods for image recognition.
Two novel evaluation protocols for generalized zero-shot learning are proposed in \cite{chao2016empirical} and \cite{xian2017zero}. However, the above generalized zero-shot learning methods are mainly for image.
Our work is evaluated under both settings with multiple evaluation metrics for generalized video zero-shot learning problems.
\section{Our Method}
\label{ourme}
According to different applications, our method can be divided into two parts. Firstly, we provide a baseline for video action recognition at both conventional zero-shot learning setting and practical generalized zero-shot learning setting. We evaluate several zero-shot leaning approaches on a standard dataset with multiple evaluation metrics.
Secondly, we take great efforts to construct a surveillance video dataset which includes a set of criminal action classes. Then we treated the whole ActivityNet dataset as seen class and our own dataset as unseen class. The knowledge learned from web video dataset is transferred to detect the actions in surveillance video. Next, we will describe both parts in detail.
\begin{figure}[t]
\begin{minipage}[b]{0.96\linewidth}
  \centering
  \centerline{\includegraphics[width=11cm,height=6cm]{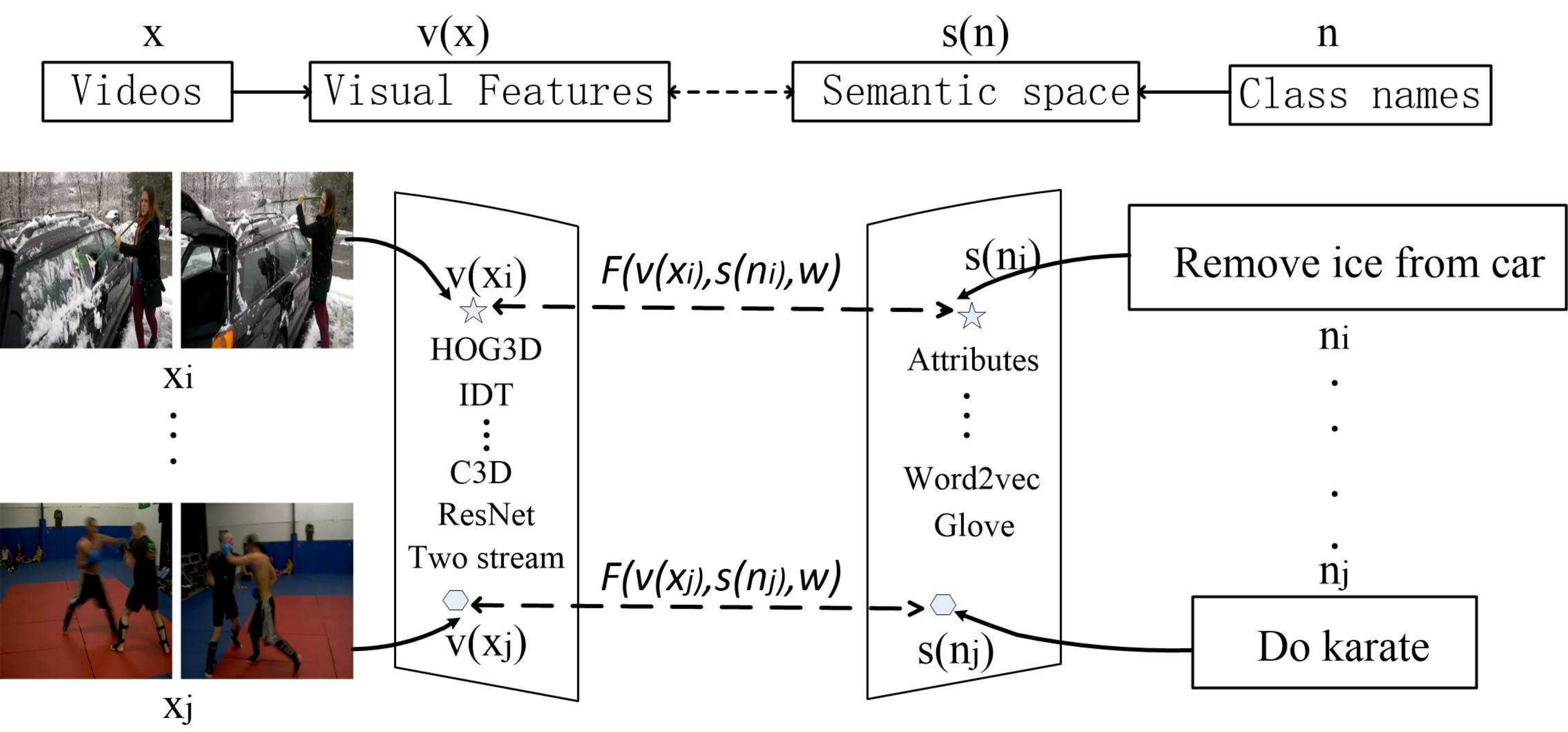}}
\end{minipage}
\caption{Demonstration of typical zero-shot learning paradigms. The solid lines represent the known relations and the dash lines represent the relationships need to be learned. The arrow direction shows their dependence relation. Typical paradigms of zero-shot learning are distinguished in the implementation for bridging images
with classes descriptions. (Best viewed in color and with zoom.)}
\label{fig:02}
\vspace{-2mm}
\end{figure}
\subsection{Zero-shot Learning.}
For the task of action recognition in surveillance video, no enough available training data results in big challenges for the current action recognition algorithms.
On the other hand, there are numerous well-constructed and large-scale video datasets which are publicly available.
Considering the above factors, zero-shot learning \cite{gan2016concepts,gan2017deck} is an attractive alternative approach to deal with this problem.%
Zero-shot learning is the special case of transfer learning where partial classes do not have any training
instance available but with some semantic descriptions of these class names. Then during the testing time, it attempts to recognize the examples of these unseen classes. The demonstration of typical zero-shot learning paradigms is shown in Fig.~\ref{fig:02}. Generally, the pipline of zero-shot learning algorithm involves three measures:

\begin{enumerate}[1)]
\item The visual space of videos is constructed by extracting either hand-crafted feature or deep feature, such as IDT \cite{wang2013action} and two-stream \cite{simonyan2014two}.
\item The class names are mapped into the label embedding via an attribute-based representation or word vector representation.
\item Bridging videos with classes of descriptions to relate the seen categories and the unseen ones.
\end{enumerate}
The key issue of zero-shot learning is to learn transferable knowledge for bridging the semantic
gap between visual instance and actions classes. In summary, the majority of recent zero-shot learning directly learns a mapping from a visual feature space to a semantic space, while other zero-shot learning approaches focus on learning non-linear multimodal embeddings.

In our work, as illustrated in Fig.~\ref{fig:01}, current web video datasets in action recognition act as training set, where each class usually provides with sufficient labeled data and also their semantic descriptions. Our own surveillance video dataset plays as testing set which only has the semantic information about the label names.
we take advantage of well-labelled web-scale video datasets to recognize the activities in surveillance video by learning sharable knowledge from the web-scale video and surveillance video.

\subsection{Generalized zero-shot learning.}
Zero-shot learning for action recognition in video has been widely studied in the computer version community.
However, to the best of our knowledge, there is no related work for action recognition in video at generalized zero-shot
learning setting, which is more realistic and practical. In order to supply a baseline for this meaningful task, we do extensive
experiments on a large scale dataset and our own dataset at both settings, namely conventional zero-shot learning setting and practical generalized zero
shot learning setting.
We use multiple metrics to evaluate classifiers from the following zero-shot learning
approaches: Convex combination of Semantic Embeddings (ConSE) \cite{ norouzi2013zero}, Latent Embeddings (LatEm) \cite{ xian2016latent}, and Synthesized Classifiers (SynC) \cite{changpinyo2016synthesized}.
Then these approaches are introduced in detail.

\begin{figure}[t]
\begin{minipage}[b]{0.96\linewidth}
  \centering
  \centerline{\includegraphics[width=11cm,height=5.5cm]{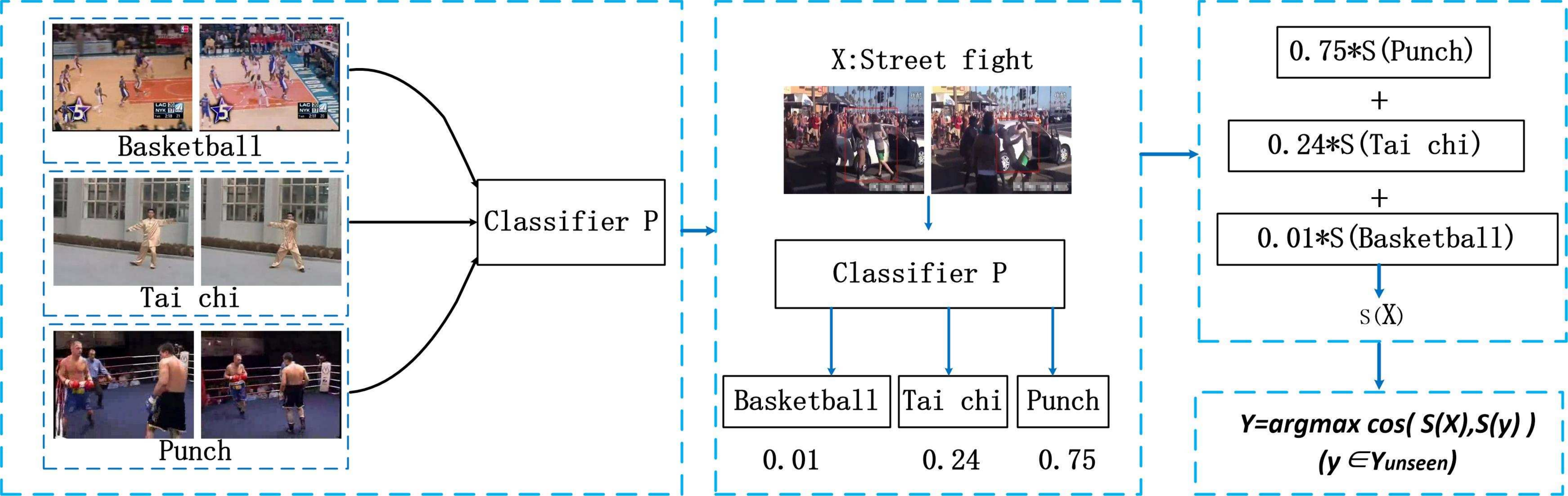}}
\end{minipage}
\caption{The demonstration of ConSE pipeline using a concrete example. S(x) is the function that projects the label names into semantic space.
Common methods include attribute based method and word vector representation. }
\label{fig:03}
\end{figure}

\subsubsection{ConSE}
ConSE \cite{ norouzi2013zero} is one of the classification based methods, which follows the traditional machine learning method
and learns a classifier from seen inputs to seen labels rather than learn a regression function explicitly.
The complete pipeline of ConSE is illustrated in Fig. \ref{fig:03} through a concrete example. ConSE \cite{ norouzi2013zero} is a two-stage approach that firstly predicts seen class
posteriors via the classifier trained on the labelled seen data. Then it projects visual feature into the semantic space by taking the convex combination of top $t$ most possibly seen classes (in this case $t=3$). Finally, the category with closest embeddings among all unseen classes in the semantic space is the inferred result.
In details, a classifier $p_{tr}$ is trained on the training set $y_{tr}$. The probability of the action class can be obtained through the follow equation:
\begin{equation}
f(x,t) = \mathop{\argmax}_{y \in y_{tr}}\ p_{tr}(y|x),
\end{equation}
where $y_{}$ denotes the $t^{th}$ most likely training label for $x$ according to $p_{tr}$. For example, $f(x,1)$ denotes the most likely training label for an image $x$ according to the classifier $p_{tr}$.
Then next step is to transfer the probabilistic predictions of the classifier beyond the training labels, that is, to a set of unseen labels.
In detail, given the top $t$ predictions for an input $x$, the trained model deterministically predicts a semantic embedding vector $s(f(x,t))$ for an input $x$.
Then we compute the $g(x)$, which is the predicted embedding of the data sample $x$ in the semantic space through
\begin{equation}
g(x)=\frac{1}{Z} \sum_{i=1}^Tp_{tr}(f(x,t)|x)\cdot s(f(x,t)),
\end{equation}
where $p_{tr}(f(x,t)|x)$ is the predict probability for each seen class $f(x,t)$; $Z$ is a normalization factor, and $T$ is a hyper-parameter controlling the maximum number
of semantic embedding vectors. This method makes use of word vector representation of class names to build the semantic embedding. In the end, we obtain the top prediction for an input $x$ from the test label set by using cosine similarity to rank the embedding vectors. More formally,
\begin{equation}
k(x,1)=\mathop{\argmax}_{y \in y_{test}} \cos(g(x),s(y)),
\end{equation}
where $k(x,t)$ denote the $t^{th}$ most likely test label predicted for $x_{}$.

\subsubsection{SynC}
SynC is one of the projection based methods, which learns a mapping  between the semantic class embedding
space and  base classifiers \cite{changpinyo2016synthesized}.
The basic classifiers are trained on labeled data from the seen classes with manifold learning and formed the model parameter space.
A set of ``phantom'' object classes are introduced to align the embedding space with model space. The ``phantom'' object classes coordinates live in both the semantic and the model space. In the model space, the training classes and phantom classes form a weighted bipartite graph. The objective is to minimize the distortion error. More formally, the distortion error is formulated as follows:
\begin{equation}
\min_{w_{c},v_{r}}\|w_{c}-\sum_{r=1}^R s_{cr}v_{r}\|_{2}^2.
\end{equation}
Semantic and model spaces are aligned by embedding real ($w_{c}$) and phantom classes ($v_{r}$) in the weighted graph ($s_{cr}$).
As a consequence, this approach can align the semantic embedding that come from external information to the model space
that concerns itself with recognizing visual features.

Two settings are considered when the approach learns the base classifiers from the training data (of the seen classes only).
Firstly, the one versus other classifiers are trained with the squared hinge loss. Alternatively, the struct loss, which is similar to Crammer-Singer multi-class SVM loss, is applied to learn other classifiers.
In this paper, we do experiments with these two different loss and donate them as $SynC^{ovso}$ and $SynC^{struct}$ respectively.
Extensive experiments on several datasets demonstrate that this approach can achieve competitive performance under both settings.

\subsubsection{LatEm}
LatEm \cite{xian2016latent} propose a novel Latent Embedding (LatEm) to encode an additional non-linearity component into
linear compatibility learning framework. LatEm is a improved version of Structured Joint Embeddings (SJE) \cite{akata2015evaluation}.
In SJE, a bi-linear compatibility function is introduced to relate visual and auxiliary information:
\begin{equation}
 F(x;y;W) = {v(x)}^T W s(y),
\end{equation}
where $v(x)$ and $s(y)$ is visual and label embedding, respectively. They are both given by various methods shown in the Fig~\ref{fig:02}.
$F(:)$ is parameterized by the mapping $W$ which is to be learned.

Compared with SJE, LatEm replaces the bilinear compatibility model with a piece-wise linear one£º
\begin{equation}
 F(x;y;W) = \max \limits_{1\leq i \leq K} {v(x)}^T W_{i} s(y),
\end{equation}
where $K$ is a hyperparameter to represents the number of latent variables,
every $W_i$ depicts a different visual characteristic of the data corresponding to typical pose and appearance.
A pairwise ranking objective formulated in the following Equation (\ref{equ:07}) is applied to obtain multiple optimized $W_{i}$ with Stochastic Gradient Descent optimizer.
\begin{equation}
\sum_{y \in Y^{tr}} \left[  \Delta(y_{n},y)+F(x_{n};y;W)-F(x_{n};y_{n};W)  \right]_{+},
\label{equ:07}
\end{equation}
where  the indicator $\Delta(y_{n},y)$ is equal to $1$ when $y_{n}=y$, otherwise $0$.

\section{Experiments Results}
\label{exper}

\subsection{Experimental Datasets}
As demonstrating zero-shot performance on small-scale datasets is not conclusive, among the most widely used datasets for action recognition, we select one large-scale popular dataset, i.e., ActivityNet. We first provide zero-shot learning results on ActivityNet and our own dataset, respectively.
Then we present results at the generalized zero-shot learning setting on both datasets.
\begin{table}[t]
\centering \caption{Comparison of statistics with human action recognition datasets. ``Actions'', specifies the number of action classes; ``Clips'', the number of clips per class in the testing set; ``Variation'', the number of clips per video, lower is better.}
\begin{tabular}{lllll}%|l|c|c|c|c|c|
  \hline
  Dataset                              & Year    & Actions        & Clips      & Variation(clip/video)  \\
  \hline
  HMDB51\cite{kuehne2013hmdb51}      &2011  &  51   &  30 & 2.0   \\
  UCF101\cite{soomro2012ucf101}      &2012  &  \textbf{101}   &  37 & 1.3   \\
  Surv5H                             &2017  &  9   &  \textbf{55} & \textbf{1}   \\
  \hline
\end{tabular}
\label{tab:01}
\end{table}

\textbf{ActivityNet.} This dataset \cite{caba2015activitynet} is a large-scale and challenging dataset which contains about ten thousand training videos, five thousand validation videos, and five thousand testing videos. The total duration of the whole dataset is longer than 600 hours. This dataset offers temporally untrimmed videos,
which means some parts of these videos have no actions. More precisely, more than half of the video hours do not contain a label among the 200 activity classes defined by the dataset.
As a result, action recognition in untrimmed videos is much more challenging than in trimmed videos. In our work, we use the same evaluation criterion as in previous work \cite{caba2015activitynet,wang2016temporal}, namely top-1 accuracy. Since there is no existing split defined on ActivityNet for zero-shot learning, we randomly split the dataset into 150 seen and 50 unseen classes.

\begin{figure}[t]
\begin{minipage}[b]{0.325\linewidth}
  \centering
  \centerline{\includegraphics[width=3.7cm,height=1.25cm]{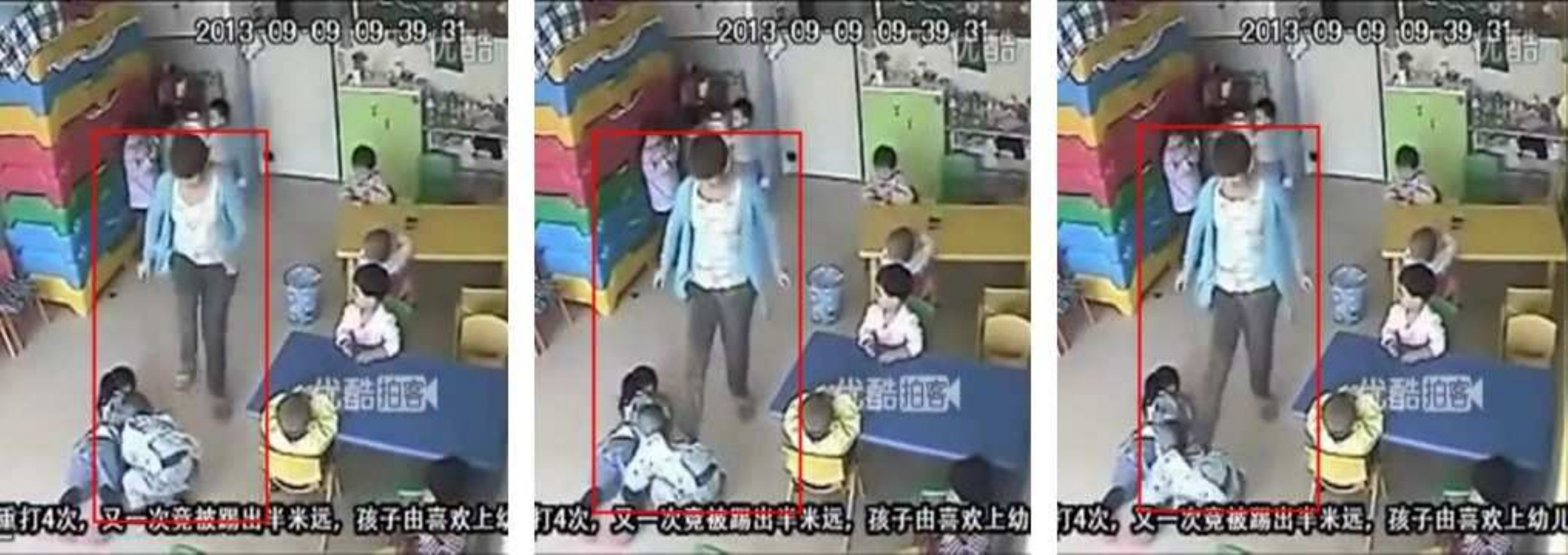}}
  \centerline{(a) Baby abuse}\medskip
\end{minipage}
\begin{minipage}[b]{0.325\linewidth}
  \centering
  \centerline{\includegraphics[width=3.7cm,height=1.25cm]{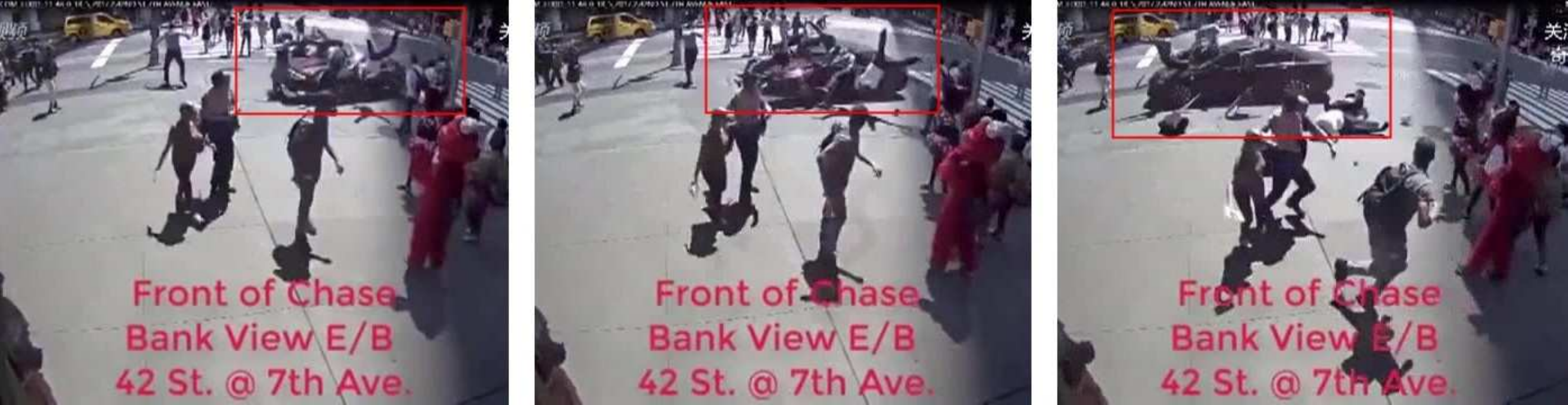}}
  \centerline{(b) Car hitting person}\medskip
\end{minipage}
\begin{minipage}[b]{0.325\linewidth}
  \centering
  \centerline{\includegraphics[width=3.7cm,height=1.25cm]{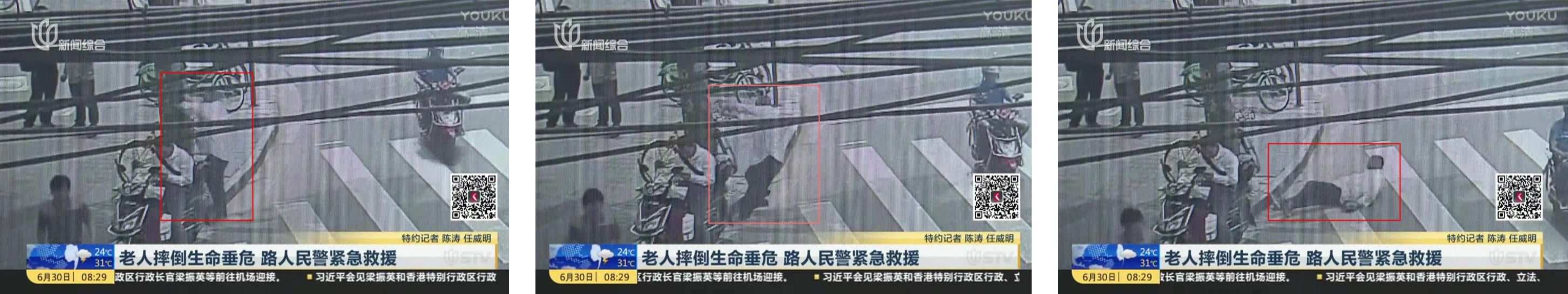}}
  \centerline{(c) Falling down}\medskip
\end{minipage}
\begin{minipage}[b]{0.325\linewidth}
  \centering
  \centerline{\includegraphics[width=3.7cm,height=1.25cm]{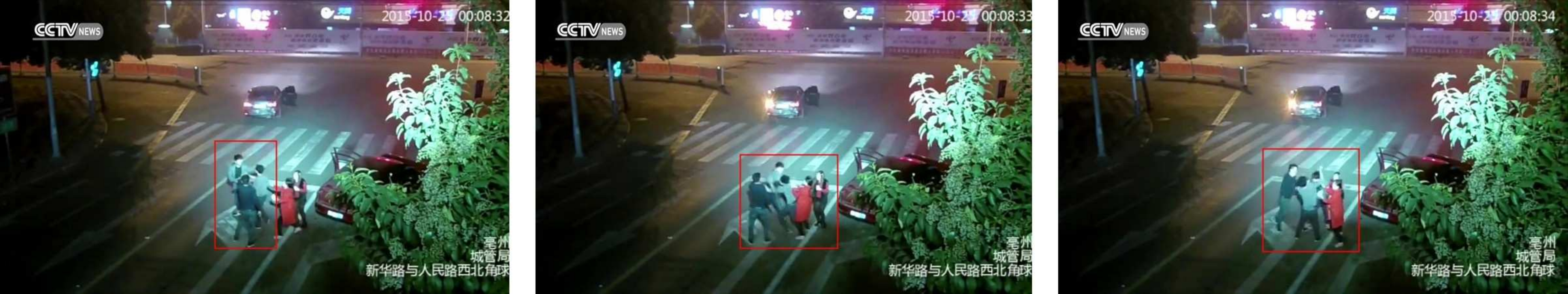}}
  \centerline{(d) Fight street}\medskip
\end{minipage}
\begin{minipage}[b]{0.325\linewidth}
  \centering
  \centerline{\includegraphics[width=3.7cm,height=1.25cm]{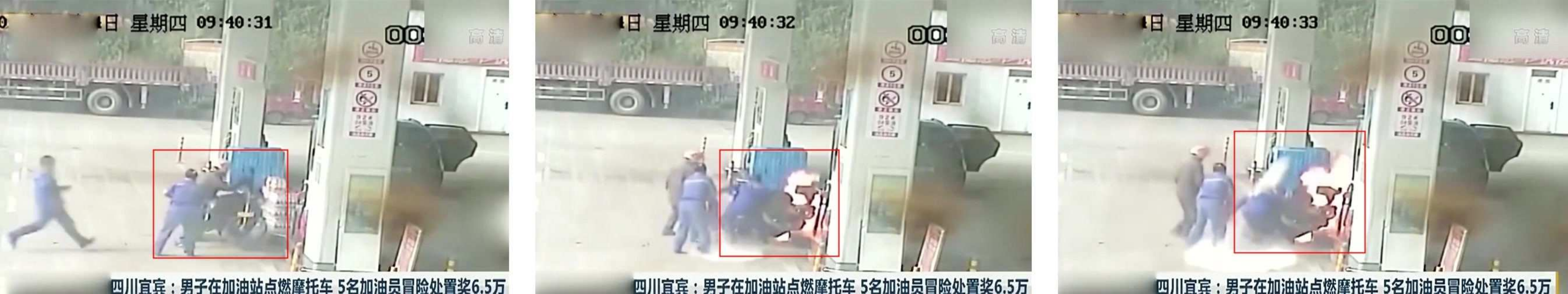}}
  \centerline{(e) Make Fire}\medskip
\end{minipage}
\begin{minipage}[b]{0.325\linewidth}
  \centering
  \centerline{\includegraphics[width=3.7cm,height=1.25cm]{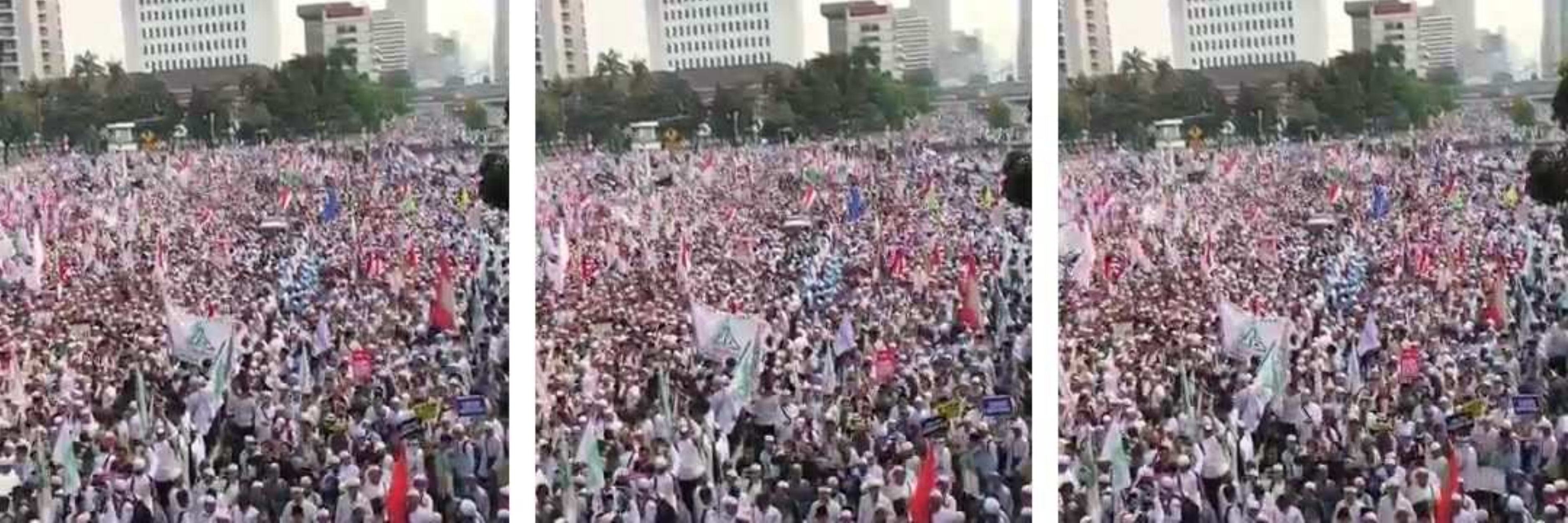}}
  \centerline{(f) Demonstrate}\medskip
\end{minipage}
\begin{minipage}[b]{0.325\linewidth}
  \centering
  \centerline{\includegraphics[width=3.7cm,height=1.25cm]{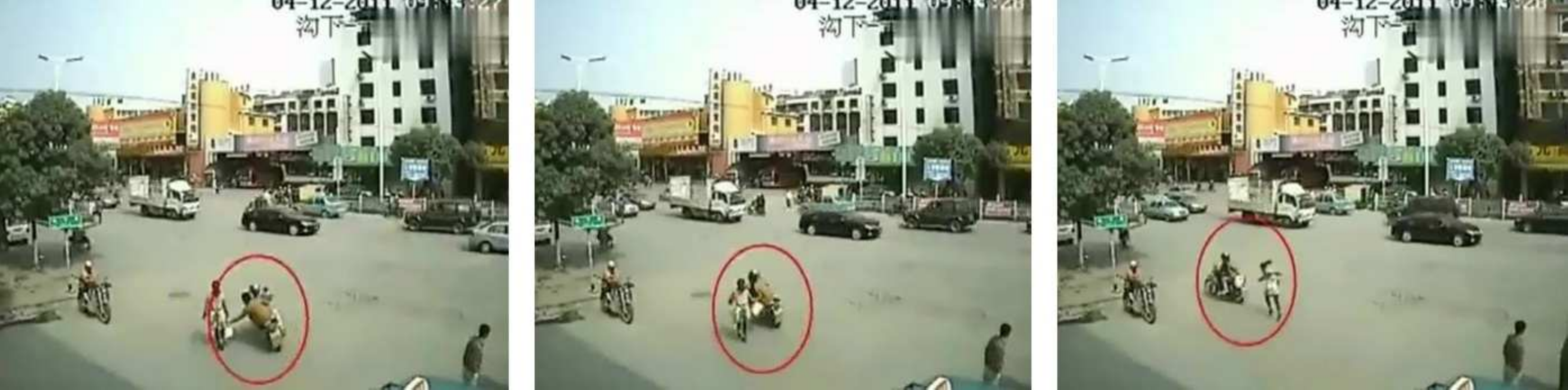}}
  \centerline{(g) Rob}\medskip
\end{minipage}
\begin{minipage}[b]{0.325\linewidth}
  \centering
  \centerline{\includegraphics[width=3.7cm,height=1.25cm]{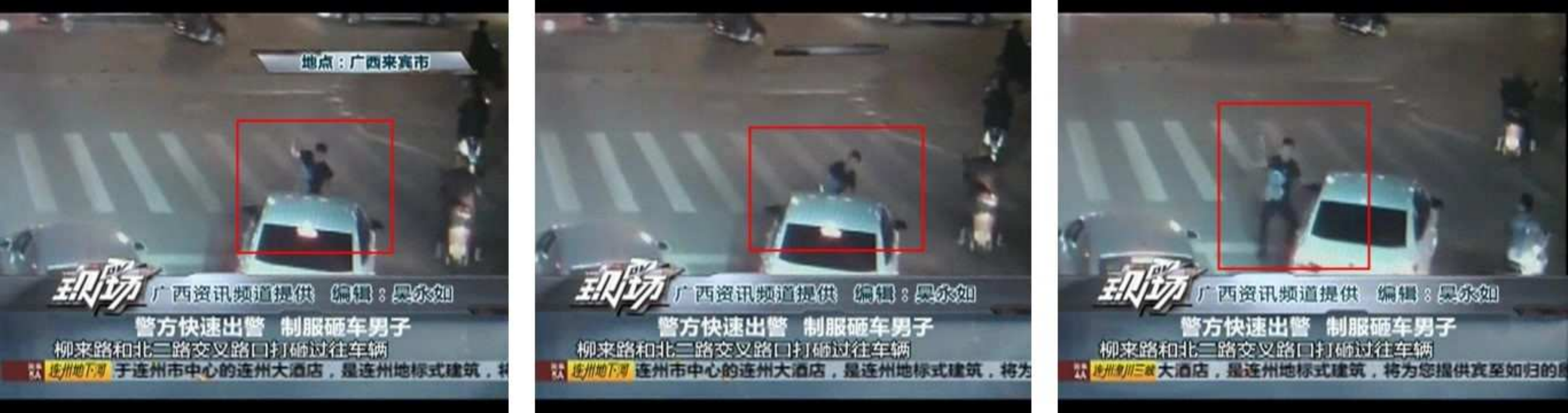}}
  \centerline{(h) Smashing car}\medskip
\end{minipage}
\begin{minipage}[b]{0.325\linewidth}
  \centering
  \centerline{\includegraphics[width=3.7cm,height=1.25cm]{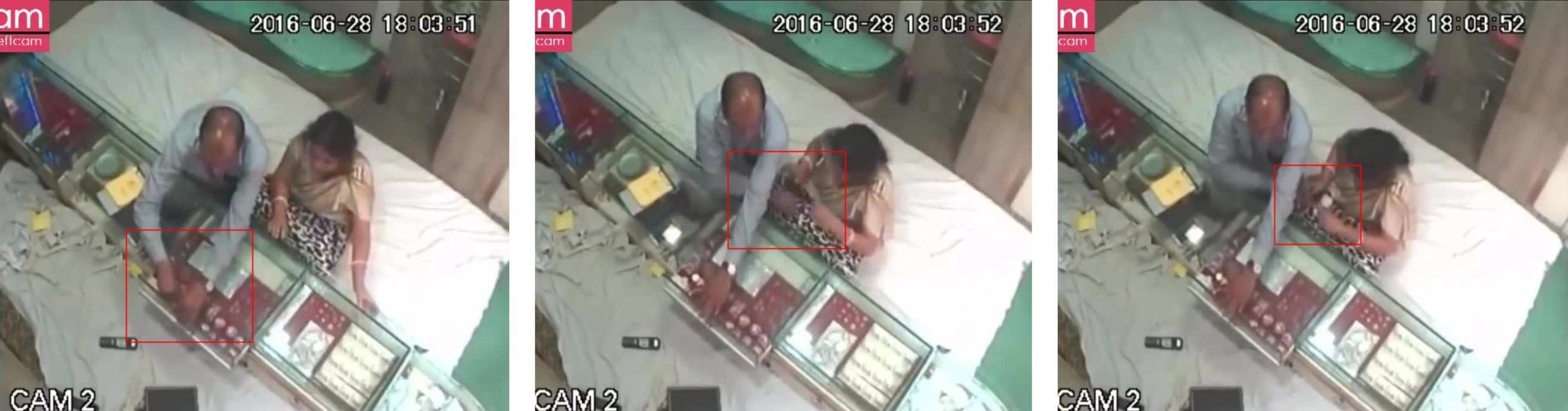}}
  \centerline{(i) Theft}\medskip
\end{minipage}
\vspace{-4mm}
\caption{Examples of nine action classes in the \emph{Surv5H} dataset. Best seen in colour and with zoom.
The action area on the picture is highlighted with a red box.
To enhance the community in the action recognition research, we will make this dataset publicly available.}
\label{fig:04}
\end{figure}

\textbf{Surv5H.} Although action recognition in surveillance video is a hot research topic in computer vision,
there is few available benchmark datasets in this area currently. To enhance the community and well evaluate the proposed method, we construct a surveillance video dataset which contains 500 videos for action recognition named Surv5H in this paper. The dataset contains nine human action classes, with at least 50 video clips for each action. Fig. \ref{fig:04} shows representative frames from all action categories.
We can conclude that recognize actions in surveillance video is more challenging than in web-scale video dataset. Because in the majority of cases a single image or several consecutive frames are not enough for recognizing the action (e.g. theft) or distinguishing classes (e.g., car hitting person vs. smashing car). The list of action classes covers: baby abuse, car hitting person, falling down, street fight, make fire, demonstrate, smashing car, robbing, and theft. Obviously, these actions are either dangerous or harmful to our lives. Consequently, it is practical to detect these actions in time and these videos of occurred rarely action are very few.
Building a list of human actions in surveillance video is challenging, as there is no public listing available with suitable visual action classes.
Therefore, we have to observe some actions that surround us and seriously influence our lives. Secondly, these videos are drawn from some video-sharing websites by matching video descriptions or titles with the actions list. The dataset is dirty due to the amount of noise and variation in the mined data. Hence, some videos are removed immediately from the pool if it does not contain an example of the action or not recorded by surveillance cameras. Further efforts have been made to find a position in all of these videos where one of the actions is potentially occurring. As a result, trimmed videos can be obtained by leveraging the point. Finally, we remove several classes that are made up of low quality or extremely few relevant candidates. \newline
\begin{figure}[t]
\begin{minipage}[b]{0.96\linewidth}
  \centering
  \centerline{\includegraphics[width=5cm,height=4.0cm]{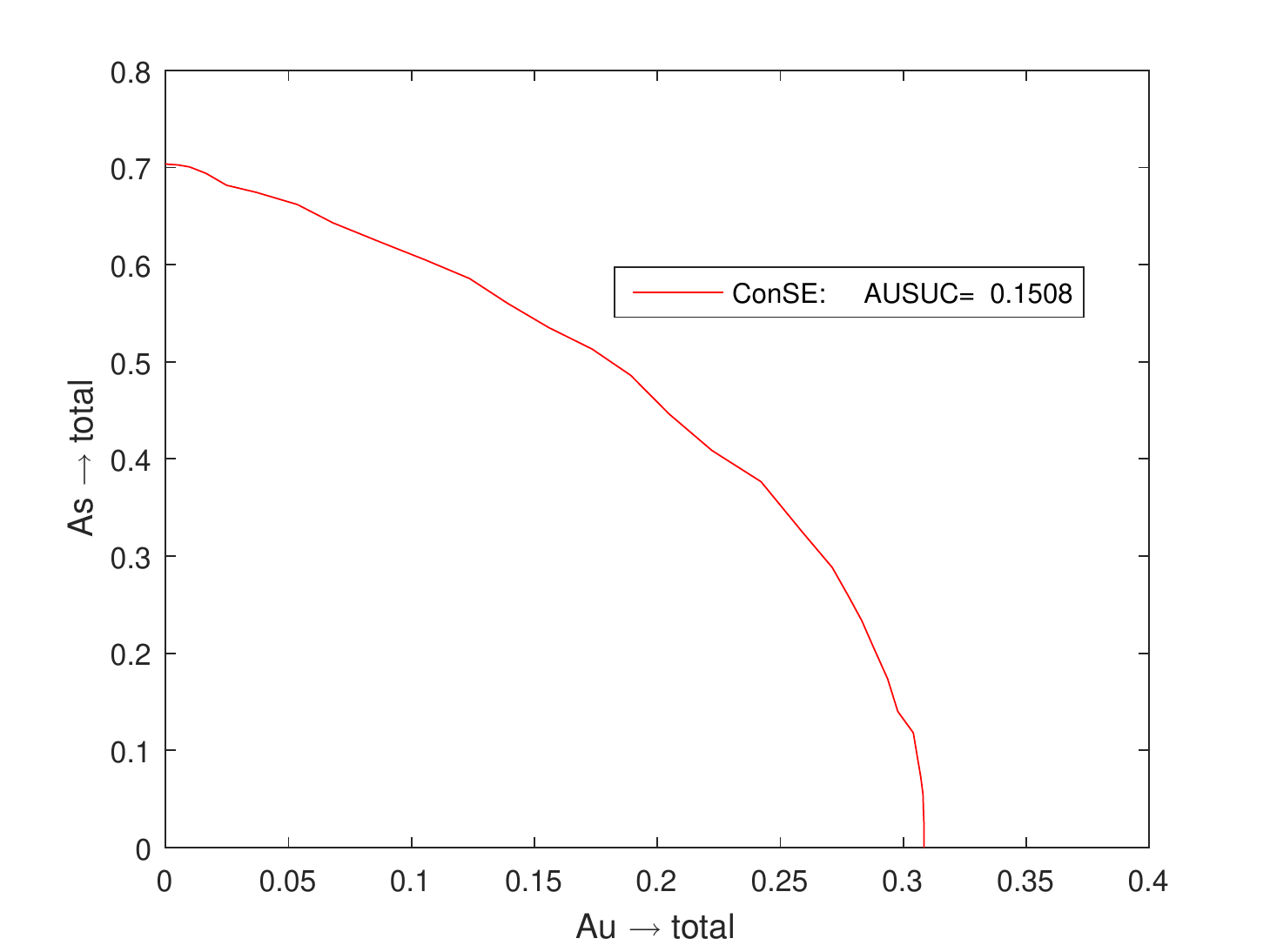}}
  \caption{The Area Under Seen-Unseen accuracy Curve (AUSUC) obtained by varying $\gamma$ in the Equation \ref{equ:06} using the method ConSE on the ActivityNet dataset.}
  \label{fig:05}
\end{minipage}
\end{figure}
\indent Note that all above processes are done by manual labeling so that we can identify that the supposed action was actually
occurring during each video. Table \ref{tab:01} compares the size of Surv5H to a number of widely used human action datasets.
In some datasets, the action in each clip was performed by same person. Our dataset is diverse as each instance
in Surv5H is from a different video. We make our dataset publicly available and anticipate its use by computer vision
researchers focused on action recognition in surveillance video.
\begin{table}[t]
\centering \caption{Classification accuracies (\%) on classic zero-shot learning ($A_{U} \rightarrow U$), multi-class classification for seen classes ($A_{S} \rightarrow S$), and generalized zero-shot learning ($A_{S} \rightarrow total$ and $A_{U} \rightarrow total$) on ActivityNet. More precisely, $A_{U}$$\rightarrow$U is the standard performance metric used for conventional zero-shot learning which indicate the accuracy of
classifying testing data from Unseen into Unseen, $A_{S}$$\rightarrow$S is  the standard metric for multi-class classification classifying testing data from Seen
into Seen, and finally $A_{S}$$\rightarrow$total and $A_{U}$$\rightarrow$total represent the accuracies of classifying testing data from either seen or unseen
classes into the joint labeling space.}
\begin{tabular}{lllllll}
  \hline
  Method                              & $A_{U}$$\rightarrow$$U$    & $A_{S}$$\rightarrow$$S$        & $A_{U}$$\rightarrow$$total$      & $A_{S}$$\rightarrow$$total$      &Mean  \\
  \hline
  ConSE\cite{ norouzi2013zero}                                &30.84  &  70.37   &  0 & 70.37  & 35.19 \\
  $SynC^{o-vs-o}$\cite{changpinyo2016synthesized}             &33.86  &  61.09   &  0 & 61.09  & 30.55   \\
  $SynC^{struct}$\cite{changpinyo2016synthesized}             &30.07  &  71.29   &  0 & 71.28  &35.64   \\
  LatEm\cite{ xian2016latent}                                 &28.51  &  65.71  &1.39 & 64.69  & 32.35 \\
  \hline
\end{tabular}
\label{tab:02}
\end{table}

\subsection{The Basic Experimental Setting}

\textbf{Visual features.} As illustrated in Fig~\ref{fig:02}, there are many alternative approaches can map the video into the visual space, such as IDT \cite{wang2013action}, C3D \cite{tran2015learning} and CNN. The computational cost of IDT is too huge to be applied in the large-scale datasets. C3D is capable of encoding the video into compact and effective feature by pre-training weight parameters on Sports-1M. However, it is likely to violate the zero-shot learning assumption since extracting video feature via a pre-trained C3D on a large dataset \cite{tran2015learning} that contains zero-shot testing classes. Meanwhile, features extracted by CNN have been proved to be the powerful information for action recognition and multimedia event recognition, which can capture the static information directly and then obtain the temporal feature of the whole video by encoding the individual frame information.

Considering the above factors, the latest extremely deep model architecture \cite{he2016deep} is introduced to encode the video information. First of all, frame-level features are extracted from the pool5 layer of ResNet-50 at one frame per second rate. The representation for the entire video is constructed by mean pooling of all the individual frames' features. L1-normalisation operator is added after mean pooling to acquire the final efficient video feature. All visual features are extracted with the Caffe \cite{jia2014caffe} package.

\textbf{Semantic spaces.} Although there is no available positive sample in zero-shot learning, we can take advantage of the relationship between seen class and unseen classes to transfer the knowledge from seen to unseen ones. In other words, classifiers for unseen classes are learned by projecting them to seen ones. This is often performed by embedding both seen and unseen classes into a common semantic space, such as word2vec representations of the class names or visual attributes. This common semantic space is able to transfer knowledge of the seen classes into the unseen ones. For example, visual semantic attribute representation for action categories is provided manually according to experts' knowledge.
However, this procedure is too labor exhaustively which also possibly introduces human bias. Moreover, Chuang et al. \cite{chuang2015exploring} points out that class-attribute representation is complicated. Hence it is hard to model the relationship between seen and unseen ones.

Considering the above factors, in this paper, we train word2vec on a large-scale text corpus to generate semantic word representations. Besides, some names of action class have no corresponding vectors with the method of word2vec due to its rare appearance. To address this issue, we replace the rarely used term with common words. For instance, ``Powerbocking'' is replaced by ``jumping stilts'' without losing the main semantic information.
To represent the semantic information of action in a fixed length, we simply average all the word vectors in the name to obtain the compact semantic feature.

\begin{table}[t]
\centering \caption{Classification accuracies(\%) on classic zero-shot learning , multi-class classification
for seen classes , and generalized zero-shot learning on Surv5H.}
\begin{tabular}{lllllll}
  \hline
  Method                              & $A_{U}$$\rightarrow$$U$    & $A_{S}$$\rightarrow$$S$        & $A_{U}$$\rightarrow$$total$      & $A_{S}$$\rightarrow$$total$      &Mean  \\
  \hline
  ConSE\cite{ norouzi2013zero}                                &23.29  &  64.41   &  0 & 62.44  & 31.22 \\
  $SynC^{o-vs-o}$\cite{changpinyo2016synthesized}             &23.45  &  59.87   &  0 & 59.87  & 29.94   \\
  $SynC^{struct}$\cite{changpinyo2016synthesized}             &23.85  &  62.44   &  0 & 62.44  & 31.22   \\
  LatEm\cite{ xian2016latent}                                 &20.75  &  59.93   &  0 & 59.93  & 29.97 \\
  \hline
\end{tabular}
\label{tab:03}
\end{table}

\textbf{Evaluation protocols.} Since action recognition is viewed as a classification problem, the evaluation metric of zero-shot learning is the common classification accuracy.
In generalized zero-shot learning setting, a reasonable performance criteria requires to take consideration of two aspects: recognizing data from seen classes versus those from unseen ones, named $A_{S} \rightarrow total$ and $A_{U} \rightarrow total$ in the Table \ref{tab:02}.
In our work, we adopt two recently proposed performance metric to characterize the trade-off between two above forces. Xian et al. \cite{xian2017zero} proposed to use harmonic mean of training and testing accuracies to examine the utility of generalized zero-shot learning methods. The formulation is presented as follows:
\begin{equation}
H = 2(Acc_{y_{train}} \times Acc_{y_{test}} )/(Acc_{y_{train}} + Acc_{y_{test}} ),
\end{equation}
where $Acc_{y_{train}}$ and $Acc_{y_{test}}$ indicates the accuracy of samples from seen ($y_{train}$) and samples from unseen ($y_{test}$) classes, respectively.
In our work, we simply adopt the mean of $Acc_{y_{train}}$ and $Acc_{y_{test}}$ to evaluate the performance due to rather small value of $Acc_{y_{test}}$.

Recently, Chao et al. \cite{chao2016empirical} propose the other performance criterion which calculates the Area Under Seen-Unseen Accuracy Curve (AUSUC) to balance two conflicting metrics. According to Table \ref{tab:02}, it is obvious that the scores of the discriminant functions for the seen classes are always greater than the scores for the unseen classes. The scores for the seen classes are reduced by introducing a parameter $\gamma$ in the formulation to exploit the balance:
\begin{equation}
 \hat{y}=\mathop{\argmax}_{c \in total} \ m_{c}(x)-\gamma \cdot \mathbb I[c \in Seen],
 \label{equ:06}
\end{equation}
where the indicator $\mathbb I[.] \in \{0,1\}$  indicates whether or not $c$ is a seen class and $\gamma$ is a adjustable parameter.
A series of classification accuracies ($A_{U} \rightarrow total$ and $A_{S} \rightarrow total$) can be computed by varying the calibration factor $\gamma$.
Fig. \ref{fig:05} plots those points for the ActivityNet using the models constructed by the algorithm
in \cite{norouzi2013zero} based on class-wise cross validation.

\begin{figure}[t]
\begin{minipage}[b]{0.48\linewidth}
  \centering
  \centerline{\includegraphics[width=5cm,height=4cm]{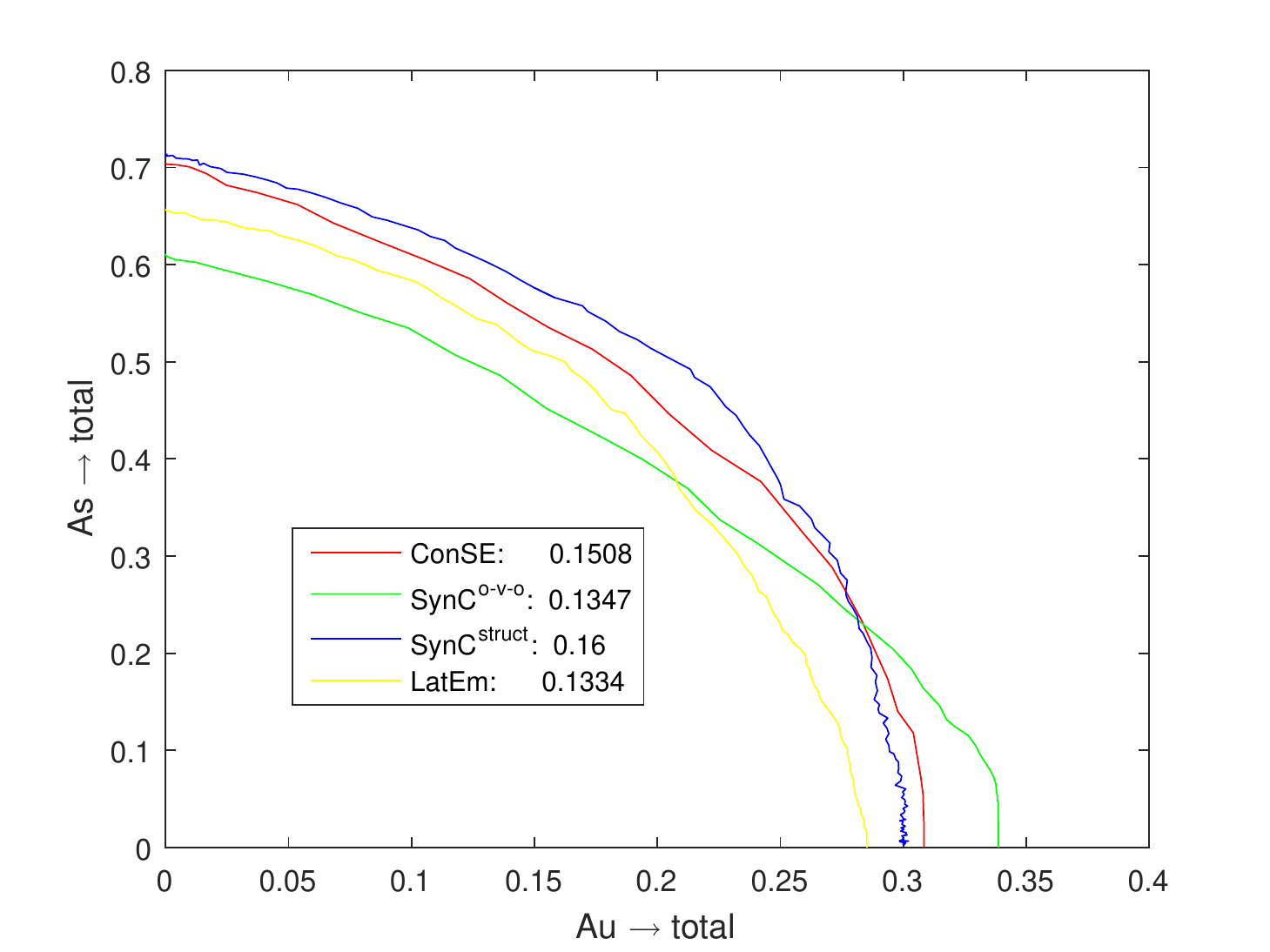}}
  \centerline{(a)\textbf{ActivityNet}}\medskip
\end{minipage}
\begin{minipage}[b]{0.48\linewidth}
  \centering
  \centerline{\includegraphics[width=5cm,height=4cm]{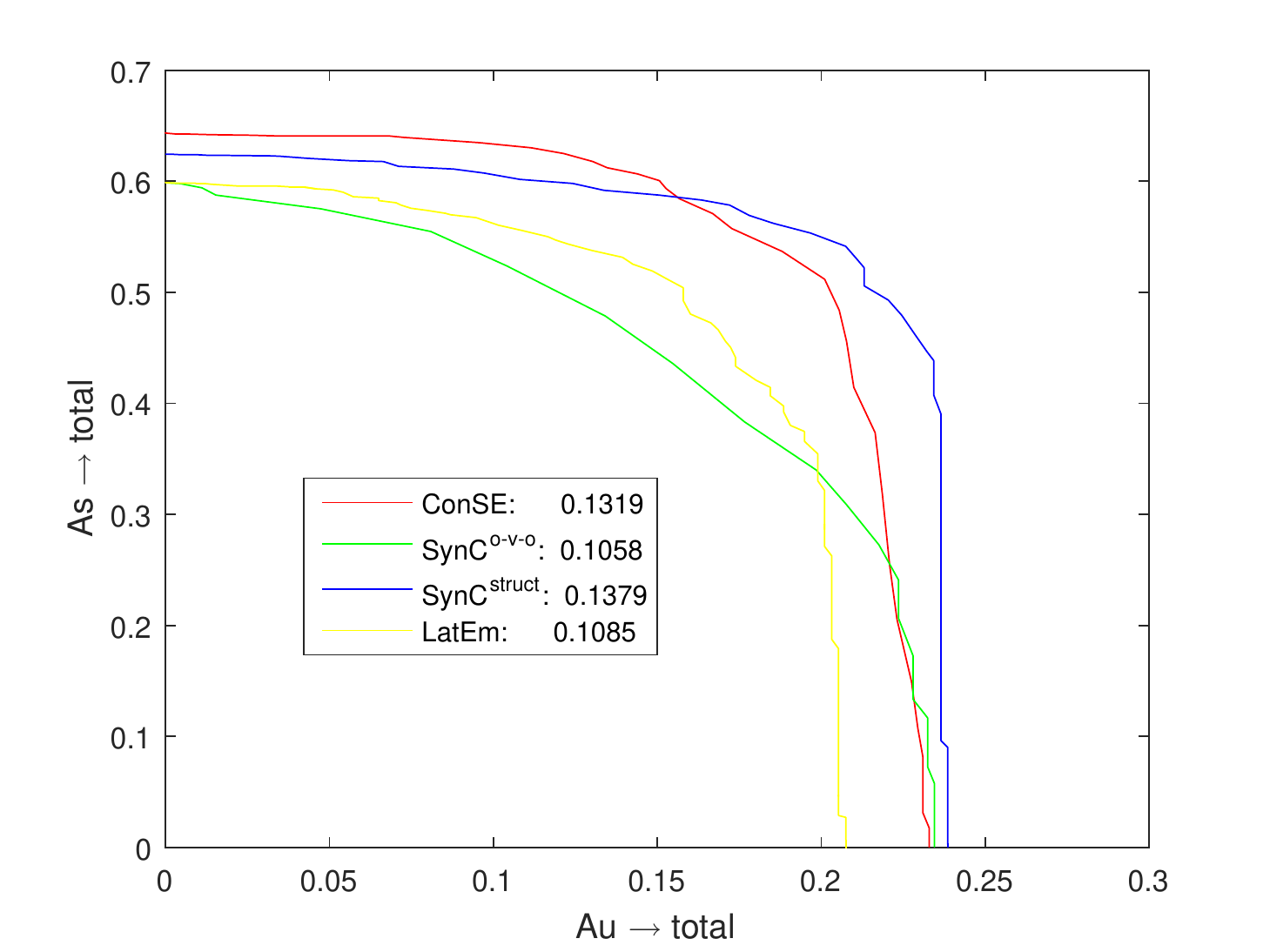}}
  \centerline{(b) \textbf{Surv5H}}\medskip
\end{minipage}
\vspace{-4mm}
\caption{Comparison between several zero-shot learning approaches on the task of generalized zero-shot learning for \textbf{ActivityNet} and \textbf{Surv5H}
using the metric of AUSUC.}
\label{fig:06}
\vspace{-4mm}
\end{figure}

When $\gamma$ = 0, the rule reverts back to the common zero-shot learning criterion which only consists of discriminant functions $m_{c}(x)$.
Besides, two extreme cases of $\gamma$ are taken into account. When $\gamma \rightarrow + \infty$, the classification rule will ignore all seen classes and classify all data points into one of the unseen classes. On the contrary, the classification rule only considers the label space of seen classes as in standard multi-way classification when $\gamma \rightarrow - \infty$. The last but not the least, classifiers with higher values of AUSUC
may be more likely to perform well in balance for the task of generalized zero-shot learning. Fig. \ref{fig:06} contrasts in detail several zero-shot learning approaches when tested on the task of generalized zero-shot learning using the evaluation protocol of AUSUC.

There are no previously established benchmark works for generalized zero-shot learning in the research of video action recognition. We thus construct the generalized zero-shot learning tasks by composing testing set as a combination of videos from both seen and unseen classes. On ActivityNet, to make our results in a larger perspective, we also perform three trial experiments by randomly splitting the training classes and testing classes. We further show the average results of three trial experiments in Table \ref{tab:02}.
Similarly, several algorithms are summarized in Table \ref{tab:03} when tested with the task of zero-shot learning and generalized zero-shot learning where the whole ActivityNet is token as seen class and Sur5H is considered as unseen data. Note that some zero-shot learning approaches may be sensitive to parameter setting, in all experiments of this paper, we therefore perform hyperparameter search on a validation set that is disjoint from training classes.

\subsection{Experiment Results Analysis}
According to above experiment results, we briefly describe and contrast them to draw the following conclusions:%Table \ref{tab:02} and Table \ref{tab:03}

Firstly, generalized zero-shot learning is very difficult. At the setting of generalized zero-shot learning, the classification accuracy for unseen classes ($A_{U}$$\rightarrow$$total$) drops significantly from the performance in
conventional zero-shot learning ($A_{U}$$\rightarrow$$U$), while that of seen ones ($A_{S}$$\rightarrow$$total$) remains roughly the same as in the multi-class task ($A_{S}$$\rightarrow$$S$). Experiment statics demonstrate that nearly all testing data from unseen classes are misclassified into the seen classes as these models are trained on the seen (training) data. In summary, the performance for generalized zero-shot learning not only relies on the ability to accurately recognize the new classes, but lies on the capacity for distinguishing the seen and unseen classes.
To ensure that our code is correct, we have run our code on the AwA dataset and obtained the same performance as \cite{chao2016empirical}.

Secondly, in terms of accuracy, the classification performance for multi-class task is always considerably better than in the zero-shot setting. It explicitly demonstrates zero-shot learning is a challenging problem either in conventional setting or generalized setting as there is no available training samples for the testing classes. On the current situation, it is obviously that we are far from the ideal performance. According to the crucial components of the typical zero-shot learning framework, we might have three candidates to achieve improvement: 1)discriminative visual representation, 2)informative semantic embedding, and 3)the bridge connecting the huge semantic gap between visual and semantic spaces. To some degree, the performance of supervised learning reflects the representational capacity of visual features. According to the accuracy of standard multi-class classification, we conclude that designing more powerful visual feature is a promising direction for the zero-shot recognition.

Thirdly, according to Fig. \ref{fig:06}, Table \ref{tab:02} and Table \ref{tab:03}, the trend among several zero-shot learning algorithms on the traditional zero-shot setting is almost the same as on the generalized setting. More specially, $SynC^{struct}$ always obtained the superior performance at both the zero-shot learning and generalized setting among the three zero-shot learning approaches. This tendency might imply that developing a powerful zero-shot learning model can also improve the performance at the generalized setting. On the contrary, a robust classifier on the generalized setting might realize the high accuracy on traditional zero-shot setting since both the conventional zero-shot learning and generalized zero-shot learning have the same goal: accurately recognize new categories without having seen the instances of these classes before.

Finally, the performance of Sur5H is always worse than these on ActivityNet. Action recognition in surveillance video is more challenging than in the web video, because some actions occurred in the dark night, which result in worse lighting conditions and motion blur in the video. Some examples are shown in the Fig~\ref{fig:04}. Besides, some surveillance videos recorded from the long-distance or high-position surveillance cameras is another huge obstacle to accurately recognize the action because of the small size of event occurrence area. Last but not the least, several actions occurred in surveillance video are extremely complex and occurred only in a moment. It is even hard for human being to accurately detect these actions, such as theft.
 %\newline
% \newline
 %\newline
% \newline
\section{Conclusion}
\label{concl}
\indent In this work, we gain insight into action recognition in surveillance video with generalized zero-shot learning method. We firstly do an extensive empirical study of several representative zero-shot leaning methods at both conventional zero-shot learning setting and generalized zero-shot learning setting on a large-scale web video dataset. To our best knowledge, this is the first work studied in video action recognition at generalized zero-shot setting.
Moreover, we construct a clean surveillance video dataset which is utilized to verify the effectiveness of our method.
Finally, we solve the issue of action recognition in surveillance video by deploying generalized zero-shot learning.
We consider the findings in this paper as a starting point for future research. In the future, we will focus our study on developing discriminative visual features and powerful semantic spaces for action recognition in surveillance videos.


\end{document}